\documentclass{article}

\usepackage{url}
\usepackage{natbib}  
\usepackage{caption} 

\usepackage{multirow}
\usepackage{tabu}
\usepackage{booktabs}
\usepackage[np,newcolumntypes]{numprint}
\npthousandsep{,}
\npproductsign{\times}
\npdecimalsign{.}

\usepackage[latin1]{inputenc}   
\usepackage[T1]{fontenc}   

\usepackage{tikz}
\def\checkmark{\tikz\fill[scale=0.4](0,.35) -- (.25,0) -- (1,.7) -- (.25,.15) -- cycle;} 

\usepackage{enumitem}
\usepackage{natbib}
\usepackage{xspace}
\usepackage{soul}
\newcommand{\fsl}{\textsl}
\usepackage{xcolor}

\newcommand{\mps}[1]{}

\newcounter{textboxno}
\usepackage[many]{tcolorbox}
\newtcolorbox{mybox}[2][]{%
boxsep=3pt,left=2pt,right=2pt,bottom=5pt,
width=\columnwidth,
boxrule=1pt,
attach boxed title to top center = {yshift=-\tcboxedtitleheight/2},
colbacktitle=white,coltitle=black,
boxed title style={size=normal,colframe=white,boxrule=0pt}, 
interior style={white},
title={\refstepcounter{textboxno}\label{#1}
Example \arabic{textboxno}: {#2}
\def\@currentlabel{\p@textboxno\thetextboxno}},
enhanced,
float,
}

\usepackage{soul}

\newcommand{\ctext}[3][RGB]{%
  \begingroup
  \definecolor{hlcolor}{#1}{#2}\sethlcolor{hlcolor}%
  \hl{#3}%
  \endgroup
}
\usepackage[normalem]{ulem}

\newcommand{\corpus}{\textsc{MeThree}\xspace}

\title{Extracting Incidents, Effects, and Requested Advice from MeToo Posts}

\author{Vaibhav Garg,
Jiaqing Yuan,
Rujie Xi
and Munindar P. Singh 
\\
vgarg3@ncsu.edu,
jyuan23@ncsu.edu,
rxi@ncsu.edu,
mpsingh@ncsu.edu
}

\begin{document}

\maketitle
\pagestyle{plain}
\thispagestyle{plain}

\begin{abstract}

\emph{Warning: This paper may contain trigger words for some readers, especially survivors of sexual harassment}.

Survivors of sexual harassment frequently share their experiences on social media, revealing their feelings and emotions and seeking advice.
We observed that on Reddit, survivors regularly share long posts that describe a combination of (i) a sexual harassment incident, (ii) its effect on the survivor, including their feelings and emotions, and (iii) the advice being sought.
We term such posts MeToo posts, even though they may not be so tagged and may appear in diverse subreddits.
A prospective helper (such as a counselor or even a casual reader) must understand a survivor's needs from such posts. But long posts can be time-consuming to read and respond to.

Accordingly, we address the problem of extracting key information from a long MeToo post. 
We develop a natural language based model to identify sentences from a post that describe any of the above three categories.
On ten-fold cross-validation of a dataset, our model achieves a macro F1 score of \np{0.82}.

In addition, we contribute \corpus, a dataset comprising \np{8947} labeled sentences extracted from Reddit posts. We apply the LIWC-22 toolkit on \corpus to understand how different language patterns in sentences of the three categories can reveal differences in emotional tone, authenticity, and other aspects.
\end{abstract}

\section{Introduction}
\label{sec:intro}

In the United States, 81\% of women and 43\% of men have reported some form of sexual harassment or assault in their lifetime.\footnote{\url{https://www.nsvrc.org/statistics}}
In 2006, Tarana Burke, an activist, coined the \emph{MeToo} phrase for survivors to share their experiences of sexual harassment.
This led to what's known as the MeToo movement, which seeks to report sexual harassment and help survivors know they are not alone.
Reddit is a popular social media platform that hosts multiple forums called subreddits (r/meToo\footnote{\url{https://www.reddit.com/r/meToo/}}, r/SexualHarassment\footnote{\url{https://www.reddit.com/r/SexualHarassment/}}, and r/sexualassault\footnote{\url{https://www.reddit.com/r/sexualassault/}}) for survivors to share their MeToo posts.

Prior studies on MeToo posts \citep{Safecity2018,SV-Tracking2020,Silence-breakers2018,Metoo-Disclosure2019} focus on classification.
For instance, \citet{Metoo-Disclosure2019} identify posts describing MeToo personal stories, \citet{Safecity2018} identify the type of sexual harassment, and \citet{SV-Tracking2020} detect the type of sexual violence. All these existing studies identify relevant MeToo posts from a massive stream of social media text.
The expectation is that a prospective helper (e.g., the concerned authority) can provide support to the survivor of identified post.
However, merely identifying relevant posts is not enough.
A prospective helper must understand (i) what happened, (ii) how sexual harassment has affected the survivor, including the feelings and emotions they are going through \cite{Metoo-Emotions2021}, and (iii) the advice that the survivor is seeking \cite{Supportseeking2016}.
 
Reddit allows a higher number of characters (40k per post) than platforms such as Twitter (250 per post). The MeToo-related subreddits too see long posts (with mean and maximum of \np{1881} and \np{33432} characters, respectively). For a prospective helper (e.g., the concerned authority), reading long posts that regularly appear on multiple subreddits \cite{Metoo-reddit-landscape2018} can be demanding and time consuming. 
To facilitate this process, we built a natural language model that extracts (from a MeToo post) sentences describing a sexual harassment incident, its effects on the survivor, and the advice requested. We describe these three sentence categories as follows:


\begin{description}
\item[Sexual harassment incident:] Sentences describing unwelcome sexual advances, sexual behavior, requests for sexual favors, verbal or physical acts of sexual nature, offensive jokes or remarks that are either sexual or based on someone's gender.\footnote{\url{https://www.eeoc.gov/sexual-harassment}}

\item[Effects on the survivor:] Survivors describe how they are affected by revealing their feelings and emotions that arise during or after the harassment incident. Examples of effects include the survivor feeling uncomfortable due to the abuser's actions, or being angry or upset due to the harassment.

\item[Requested advice:] Sentences in which survivors seek advice from other platform users. Some examples of advice include asking if the survivor's experience is harassment, how to pursue a legal case, and how to confront the abuser.
\end{description}


\begin{mybox}[box:sentences_example]{Incident, effects, and request for advice}
\textbf{Categories:} \ctext[RGB]{255,200,255}{Sexual harassment incident}, \ctext[RGB]{200,128,0}{Effects}, \ctext[RGB]{97,189,200}{Requested advice} \bigskip 

At \emph{<job-location>}, I was appointed as \emph{<job-title>} a month ago.
\ctext[RGB]{255,200,255}{In my office, this one \emph{<person>} pats my 
shoulder and I feel his hand has lingered a little too long a couple times.} \ctext[RGB]{200,128,0}{Because of my big history around sexual harassment, I feel extremely uncomfortable with his behavior.}
\ctext[RGB]{97,189,200}{I keep thinking if I am considering his behavior inappropriate because of my history?} I understand that he wants to be friendly and build rapport, but my body thinks his behavior is little off. \ctext[RGB]{97,189,200}{Reddit, am I overthinking?}

\bigskip
\textbf{Extracted sentences}
\begin{itemize}
\item \fsl{In my office, this one \emph{<person>} pats my 
shoulder and I feel his hand has lingered a little too long a couple times.}
\item \fsl{Because of my big history around sexual harassment, I feel extremely uncomfortable with his behavior.}
\item \fsl{I keep thinking if I am considering his behavior inappropriate because of my history?}
\item \fsl{Reddit, am I overthinking?}
\end{itemize}
\end{mybox}

Example~\ref{box:sentences_example} shows a MeToo post\footnote{Due to space limitations, we have shown a short MeToo post.} and the three categories of sentences that we extract from it.\footnote{The extremely personal MeToo post is paraphrased so that it's not identifiable or searchable.} The extracted text describes inappropriate touching and the survivor's uncomfortable feeling. Moreover, it reveals that the survivor is confused and asks if they are overthinking the incident. 

In other cases, survivors may ask for advice such as how to report harassment, how to deal with trauma, and so on. Prior research \citep{Metoo-Emotions2021, Supportseeking2016} shows that it's important to understand and address the effects and the requested advice.

\subsection{Research Questions}

Accordingly, we address the following research questions.

\begin{description}
\item[RQ\textsubscript{extract}:] How can we extract sentences describing the harassment incident, its effects on the survivor, and the requested advice from a MeToo post?

\end{description} 

RQ\textsubscript{extract} is important because automatically extracting the three categories of sentences will help a prospective helper understand the incident, the effects on the survivor, and the requested advice without having to read the whole post. Hence, the prospective helper (e.g., the concerned authority) may timely address the survivor and provide some advice or support.
Traditional text summarization models are trained or evaluated on specific tasks but not on MeToo corpora \citep{SwapNet2018,Cheng-lapata-neural-2016,See-etal-2017,Aspectsummarizer2011,Coherentsummary2012}.


\begin{description}
\item[RQ\textsubscript{psycholinguistic}:] How do sentences in three categories differ in psychological aspects?
\end{description} 

While writing sentences of different categories, the survivor may choose different set of words representing distinct language patterns.
RQ\textsubscript{psycholinguistic} is important to understand how such patterns can reveal psychological aspects such as emotional tone, authenticity, and type of emotion.

\subsection{Contributions and Novelty}

We make the following contributions.

\begin{itemize}

\item To address both questions, we curate \corpus, a dataset containing \np{8947} sentences, labeled for the three categories. Constructing a sufficiently natural and precise dataset turns out to be nontrivial. We leverage active learning for labeling with tractable manual effort. 

\item To address RQ\textsubscript{extract}, we train a natural language model to extract these three categories of sentences from long MeToo posts. Our approach incorporates modern Natural Language Processing (NLP) techniques to achieve strong results.

\item To address RQ\textsubscript{psycholinguistic}, we apply the LIWC-22 toolkit \cite{LIWC2022} on \corpus, analyze aspects (such as emotional tone, authenticity, type of emotion) for three sentence categories, and compare their LIWC-22 scores (Section~\ref{sec:psycholinguistic}). 
This analysis provides a psychological understanding of the essential parts of MeToo posts. 

\end{itemize}

Existing summarization tools are domain-specific and can't be applied on the MeToo corpora. To the best of our knowledge, we are the first ones to study sentence level extraction from long MeToo posts. Moreover, we conduct a comparative study for sentences of the three categories (not done before), based on psychological aspects.

\subsection{Key Findings} 
Our model for extracting three categories of sentences yields a macro F1 score of 82\%.

Our psycholinguistic analysis on \corpus reveals the following: (1) sentences describing effects are more negative and emotional than sentences describing incidents and requested advice, (2) the requested advice sentences express a more positive tone than the sentences in the other two categories, and (3) within the effects category, anxiety is prominent, followed by sadness and positive emotion.

A small qualitative study provides additional validation for our contributions (Section~\ref{sec:outputanalysis}).
For \np{17} of \np{20} randomly selected MeToo posts, the extracted text is coherent to understand incident, effects, and requested advice. For \np{16} of \np{17} posts, we can construct a helpful response without missing out on any crucial information about the survivor's situation.

\section{The \corpus Dataset and Classifier}
\label{sec:MeThree}
We consider the problem of extracting three categories of sentences as multilabel classification task. From a post, the sentences predicted as any of the three categories are extracted.

In our adopted active learning approach, the preparation of the dataset and the development of a classifier happen hand-in-hand.
For classification, we follow a pool-based active learning approach which is known for training robust models while reducing effort on manual labeling \cite{Activelearningtheory2014}. 
We curate \corpus, a dataset comprising \np{8947} labeled sentences (from subreddits: r/meToo, r/sexualassault, and r/SexualHarassment), and train an XLNet model on \corpus.

\begin{figure}[!htb]
    \centering\includegraphics[width=0.7\columnwidth]{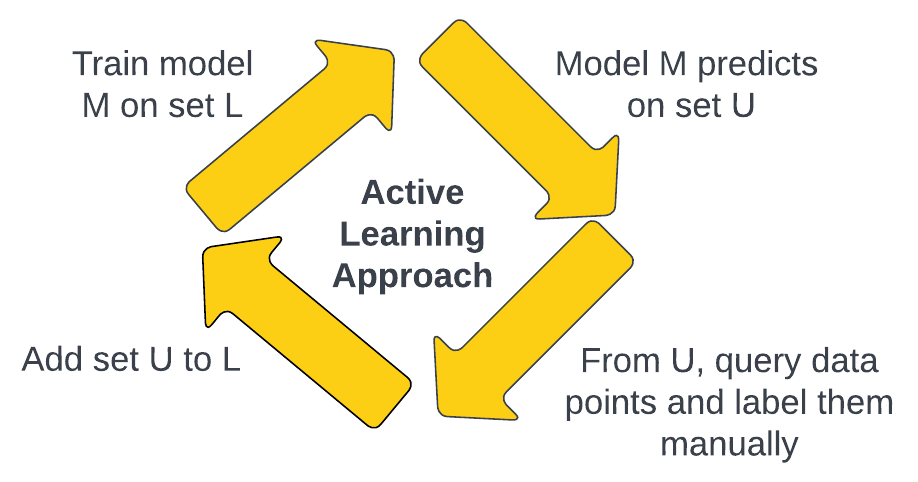}
    \caption{Active learning involves four iterative steps.}
    \label{fig:alcycle}
\end{figure}

Pool-based active learning \cite{Activelearning2012} starts with an initial dataset (denoted by L)
that we curated by selecting and labeling sentences, most of which contain certain keywords (Section~\ref{sec:labeleddata}). After curating L, in the active learning process, four steps shown in Figure~\ref{fig:alcycle} are followed and repeated multiple times. First, a model (denoted by M) is trained on the set L. To do so, we compared the performance of multiple models on the curated L and chose the best-performing one as model M (Section~\ref{sec:approach}). Second, an unlabeled dataset U is labeled by the predictions of the trained model M. In our case, since most of the sentences in L contained certain keywords, to avoid bias, we selected U from sentences without those keywords and labeled U through M's predictions. Third, from U, data points whose risk of being mispredicted is sufficiently high are queried (using a query method) and labeled manually. Fourth, U is added to L. For the last two steps, we queried misclassified sentences from the set U and labeled them manually (Section~\ref{sec:remainingal}). We also selected an appropriate query method for our approach (Section~\ref{sec:querying}). We repeated the active learning cycle five times to curate the final dataset of \np{8947} labeled sentences. We called this dataset as \corpus. In the end, we trained the final model on \corpus to extract sentences (from a long post) that are classified as the incident, its effects, and the requested advice.


\subsection{Initial Training Data for Active Learning}
\label{sec:labeleddata}
To curate our initial training data (set L) for the active learning approach, we followed four steps. First, we scraped MeToo posts from subreddits. Second, we filtered relevant posts from them. Third, we found candidate sentences for each category. Fourth, we labeled a sample of candidate sentences along with other sentences.

\subsubsection{Collecting MeToo Posts}
\label{sec:collectposts}
We scraped MeToo posts from three subreddits: r/meToo, r/sexualassault, and r/SexualHarassment,  for the period 2016-01-01 to 2021-07-18, using Reddit's Pushshift API.\footnote{\url{https://psaw.readthedocs.io/en/latest/}} In this process, we collected a total of \np{9140} posts. Of these \np{9140} posts, there were \np{263} posts whose content was deleted by the time of our scraping. That's how we were left with \np{8877} MeToo posts.

\subsubsection{Filtering Relevant MeToo Posts}
\label{sec:filterrelevant}
Some MeToo posts don't share survivors' experiences but share news articles, seek opinions about allegations against celebrities, or promote other platforms. Such posts are irrelevant to our study. We applied the following heuristics to focus on posts containing survivors' personal experiences:

\begin{itemize}
\item First-person pronouns: Many survivors while describing their personal experiences, use first-person pronouns in the title of the post. For example, ``\textbf{I} started to do something about \textbf{my} past assault, but instead of feeling better, it actually gets worse'' and ``\textbf{My} mom's boyfriend tried to get \textbf{me} to do things to him''. Thus, we checked the presence of first-person pronouns: \fsl{i}, \fsl{me}, \fsl{my}, and \fsl{mine} in the title to extract relevant MeToo posts.

\item Advice-related keywords: We observed that survivors also use advice-related keywords in the title. For example, ``Need \textbf{advice}, or support" and ``pls someone read this and \textbf{help} me figure out if i was assaulted or not". We used the keyword, \fsl{advice}, as seed and queried its synonyms from the Oxford dictionary. We obtained 25 synonyms and manually filtered four of them based on their relevance to our problem. The final list contained \fsl{help}, \fsl{suggestion}, \fsl{advice}, \fsl{guide}, and \fsl{counsel}. We referred to this list of keywords as \emph{advice keywords}. To filter relevant posts, we checked the presence of these keywords in the title. For extracting synonyms, we also explored other corpora such as WordNet \cite{Wordnet1995} but did not find synonyms that were commonly used.

\item Advice-related questions: We observed that many relevant posts ask a question (related to sexual harassment) in the title. For example, ``Was this rape?" and ``Is this sexual harassment?''. Such questions are seeking advice without mentioning any of the advice keywords. Using Part-Of-Speech (POS) tagging \cite{POS2011}, the titles that have an interrogation form and include \fsl{rape}, \fsl{harassment}, \fsl{assault}, and \fsl{abuse} as the object, were filtered.

\end{itemize}

Posts with titles satisfying one of the above rules were filtered out. We checked random \np{50} filtered posts for relevancy. Of \np{50} posts, \np{47} (94\%) were relevant because they either sought support or advice related to their case of sexual harassment. Among these \np{47} posts, we also found one post written by the survivor's friend but the post still expressed the effects on the survivor and sought advice.

In total, we obtained \np{4933} relevant posts using the above heuristics. We might have missed some relevant posts, but the objective here is to filter posts with high precision. This is because high precision (94\% in our case) means we can build a dataset of relevant sentences without further pruning. Similar heuristics are used in other studies too \cite{SV-Tracking2020}. In this study, out of \np{4933} filtered posts, 74.29\% (\np{3665}) are from r/sexualassault, followed by r/meToo (17.23\%; \np{850}) and r/SexualHarassment (8.47\%; \np{418}).

\subsubsection{Finding Candidate Sentences}
\label{sec:candidatesentences}

We split each of the \np{4933} relevant posts into sentences, using sentence tokenizer of Natural Language Toolkit (NLTK) library\footnote{\url{https://www.nltk.org/api/nltk.tokenize.html}}. That's how we obtained \np{102204} sentences. However, a random sample of these sentences was inefficient in getting sentences that describe incidents, or its effects, or requested advice. Thus, we first found candidate sentences of each category using following keywords:

\begin{itemize}
\item Sexual harassment incident: \citet{SV-Tracking2020} create a list of 27 MeToo-related verbs (such as \fsl{molest}, \fsl{touch}, \fsl{rape}, \fsl{masturbate}). We expanded the list by querying synonyms of these verbs through the Oxford dictionary. The resulting list contained \np{652} verbs. We manually checked them and found \np{539} relevant verbs, of which only 313 were unique. We called this final set of 313 verbs as \emph{harassment keywords}. We identified candidate sentences by the presence of one or more harassment keywords in them. That's how we found \np{30927} candidate sentences for the incident category.

\item Effects on the survivor: For identifying candidate sentences in this category, we leveraged two types of keywords. First, we leveraged the NRC word emotion lexicon \citep{NRC2013, NRC2010}, which contained a list of words associated with eight emotions. We considered four emotions: \fsl{anger}, \fsl{disgust}, \fsl{fear}, and \fsl{sadness}, that a survivor can express, and use lexicons associated with them. In this process, we found \np{37271} emotional candidate sentences. Second, we leveraged synonyms of the word, \fsl{feel}, that are extracted from the Oxford dictionary. We identified 14 synonyms, out of which, eight were relevant to the survivor's feelings. We referred to the final set of keywords (\fsl{feel}, \fsl{perceive}, \fsl{sense}, \fsl{experience}, \fsl{undergo}, \fsl{bear}, \fsl{endure}, \fsl{suffer}) as \emph{feel keywords}. We found \np{8617} candidate sentences containing one or more feel keywords.

\item Requested advice: We observed that many questions in MeToo posts are advice seeking. For example, ``Was it actually just a mistake and should I forgive him?'' and ``Am I blowing it out of proportion?''. Hence, we considered all questions as candidates for advice seeking sentences. We found \np{6354} such candidates. Moreover, we leveraged advice keywords to find an additional \np{2678} candidates.
\end{itemize}

We extracted synonyms from the Oxford dictionary. To find such synonyms, we tried corpora such as WordNet \cite{Wordnet1995} and PyDictionary\footnote{\url{https://pypi.org/project/PyDictionary/}} but did not find many keywords. For example, while creating the feel keywords, PyDictionary produced no synonyms, and WordNet produced one word, \fsl{palpate}, which was uncommon to describe feelings. Thus, we leveraged the Oxford dictionary to extract relevant and commonly used keywords.


\subsubsection{Labeling Sentences}
Due to presence of keywords (such as harassment keywords, feel keywords, and so on), the candidate sentences are likely to be relevant to the three categories. However, only including candidate sentences can make the training set (set L) biased toward the chosen keywords. Thus, for labeling at this step, we included random
\np{500} sentences not having any keywords, along with \np{6900} sampled candidate sentences (including sentences from all sources: harassment keywords, feel keywords, and so on). After discarding duplicates, we were left with \np{5947} sentences. 

Since a majority of \np{5947} sentences still contained chosen keywords, labeling them could still form a biased dataset. Note that this was only the initial training data (set L) in the active learning approach. Later, to mitigate bias, we kept including sentences without any keywords (set U) through multiple repetitions of active learning cycle, as described in Section~\ref{sec:remainingal}.

For \np{5947} sentences, three of the authors of this paper were the annotators. Before labeling, they were aware of the uncomfortable and disturbing text present in these sentences. For each sentence, the annotators were asked the following questions:
\begin{enumerate}
\item Does this sentence describe a sexual harassment incident? 
\item Does this sentence describe the effects of the incident on the survivor? 
\item Does this sentence ask for any advice?
\end{enumerate}

The annotators read each sentence and answered the above questions as either yes or no. Initially, two annotators labeled \np{200} sentences as per their understanding of the problem statement. Later, they discussed their disagreements and defined the final labeling instructions for all the annotators to follow. The final labeling instructions are described below:

\begin{enumerate}
\item Sexual harassment incident: We followed the definition given by the United States Equal Employment Opportunity Commission (EEOC).\footnote{\url{https://www.eeoc.gov/sexual-harassment}} Any unwelcome sexual advances, sexual behavior, requests for sexual favors, verbal or physical acts of sexual nature, offensive jokes, or remarks that were either sexual or based on someone's gender were labeled as sexual harassment. Sexual harassment is not limited to, and we considered harassment cases with all genders.


\item Effects on the survivor: We considered survivors' feelings and emotions that arose during or after the incident. Examples range from feeling uncomfortable (due to the abuser's actions) to being afraid (emotion: fear) of reporting sexual harassment. 

\item Requested advice: We considered sentences in which survivors asked for suggestions on topics related to harassment, e.g., whether to report the incident, where to get therapy from, and how to face the abuser again.
\end{enumerate}

\begin{table}[!htb]
    \centering
    \caption{Relevant examples according to labeling instructions.}
    \label{tab:sentencetype}    
    \begin{tabular}{p{5cm} c c c}
    \toprule
        Sentence & Incident& Effects & Requested advice \\
        \midrule
        \emph{\ldots he slid his hand up my leg and into my shorts.} & \checkmark & & \\
        \midrule
        \emph{\ldots I was sexually used by <abuser> on many occasions \ldots I am in a constant battle with major depression, crippling real event OCD (I ruminate for 16 hours/day) \& debilitating anxiety.} & \checkmark & \checkmark & \\
        \midrule
        \emph{\ldots I'm freaking out and have no one to talk to because no one knows about him or what happened \ldots What do I do?} &  & \checkmark&\checkmark \\
        \midrule
        \emph{Does anyone know how a legal advocate works and what you experienced with them?} & & & \checkmark \\

    \bottomrule
    \end{tabular}
\end{table}

Table~\ref{tab:sentencetype} illustrates examples of each category.\footnote{For anonymity, we have masked abusers' details.} The first example describes inappropriate physical behavior and is considered sexual harassment. The second example describes that the survivor is sexually exploited (sexual harassment) and suffers from depression and anxiety (effects). In the third example, the survivor expresses fear (by mentioning ``freak out'') and seeks advice about dealing with it. In the last example, the survivor seeks advice relating to the legal process.

All \np{5947} sentences were divided among the three annotators (let's denote them by A\textsubscript{1}, A\textsubscript{2}, and A\textsubscript{3}) such that two of the annotators labeled each sentence.  After labeling all the sentences, we obtained Cohen's kappa scores \cite{Cohen1960} of \np{0.772} (for sexual harassment incident), \np{0.774} (for effects), and \np{0.865} (for requested advice). These scores indicated that we achieved substantial agreement for two categories: sexual harassment incident and effects, and almost perfect agreement for the requested advice category. Table~\ref{tab:kappa} also shows Cohen's kappa scores for each pair of annotators. Moreover, the first author resolved all the disagreements. The labeled \np{5947} sentences form the initial training data (set L) for active learning. 

\begin{table}[!htb]
\centering
\caption{Cohen's kappa scores for each pair of annotators.}
\label{tab:kappa}
\begin{tabular}{lrrr}
\toprule
Annotators &Incident& Effects & Requested advice \\\midrule
A\textsubscript{1}, A\textsubscript{2} & 0.798 & 0.793 & 0.891 \\
A\textsubscript{2}, A\textsubscript{3} & 0.720 & 0.725 & 0.843 \\
A\textsubscript{3}, A\textsubscript{1} & 0.795 & 0.801 & 0.861 \\
\midrule
Total & 0.772 & 0.774 & 0.865\\
\bottomrule
\end{tabular}
\end{table}

\subsection{Initial Model to Extract Sentences}
\label{sec:approach}
After set L is curated, the next step is to train model M. We consider our problem as a multilabel classification task in which each sentence is an input to the model and the output has three binary labels (one label for each category). We trained and evaluated multiple methods on \np{5947} labeled sentences (set L) as described below.

For each of \np{5947} sentences, we computed embeddings such as Sentence-BERT \cite{Sentence-bert-2019}, TF-IDF \cite{Tfidf2021}, GloVe,\footnote{We used Stanford's GloVe model trained on the Wikipedia dataset, which returns a 100-dimension word vector.} \cite{Glove2014} Word2Vec\footnote{We used Word2Vec trained on the Google News dataset and returns a 300-dimension vector.} \cite{Word2Vec2013}, and Universal Sentence Encoder (USE) \cite{UniversalSentenceEncoder2018}. For each embedding, the sentence vector was used as an input to a multilabel classifier. For GloVe and Word2Vec, we averaged word vectors to form the sentence vector. For classification, we tried Logistic Regression (LR) \cite{LR2002}, Support Vector Machine (SVM) \cite{SVM2020}, and Random Forest (RF), and report the best method.


In addition to embedding-based methods, we also applied transformer-based approaches such as RoBERTa \cite{Roberta2019} and XLNet \cite{Xlnet2019}. We fine-tuned RoBERTa and XLNet on set L by adding an output layer in the forward direction. The output layer contained three units, one dedicated to each category. Both the models minimized binary cross entropy over five epochs. Moreover, the training batch size and tokenizer length were set to 32 and 256, respectively.

We computed average F1, precision, and recall scores for the approaches described above over ten-folds of set L. We also test keywords approach in which we search sentences by category-wise keywords (keywords used in Section~\ref{sec:candidatesentences}). The sentences containing keywords were predicted 1 for that category and others were predicted 0. 

TF-IDF, GloVe, Word2Vec, Keyword search, and USE underperform as compared to other methods. 

Sentence-BERT followed by SVM achieves the highest macro precision (\np{0.84}). However, it shows lower macro recall (\np{0.66}) than RoBERTa (\np{0.84}) and XLNet (\np{0.87}). Overall, XlNet outperforms all other methods by achieving the highest macro F1 score (\np{0.82}). Thus, we choose XLNet as our active learning model (model M).

\subsection{Completing Active Learning Cycles}
\label{sec:remainingal}
After model M is trained, it's time to make predictions on the set U and label it. To mitigate the risk of a biased dataset, we chose the set U to be a random sample of \np{500} sentences not containing any keywords. The already trained model M labeled set U through its predictions. From U, we queried potentially misclassified sentences for manual labeling, using a query method described in Section~\ref{sec:querying}. Further, the first active learning cycle (Figure~\ref{fig:alcycle}) was completed by adding labeled U to L. We repeated this for four more cycles that involves the training XLNet on the new L, predicting on new U, labeling new U (through M's predictions and manually labeling the queried sentences), and adding new U to L. Overall, a total of five cycles 
added total 2500 labeled sentences (each time U having 500 sentences without keywords) to the initially \np{5947} labeled ones. As a result, the final L became to be of size \np{8447}. Moreover, while selecting appropriate query method for our approach, as discussed in Section~\ref{sec:querying}, we labeled additional \np{500} sentences without keywords. By including all these labeled sentences, we formed the final dataset, \corpus, of size \np{8947}. 

\begin{figure}[!htb]
    \noindent\includegraphics[width=\columnwidth]{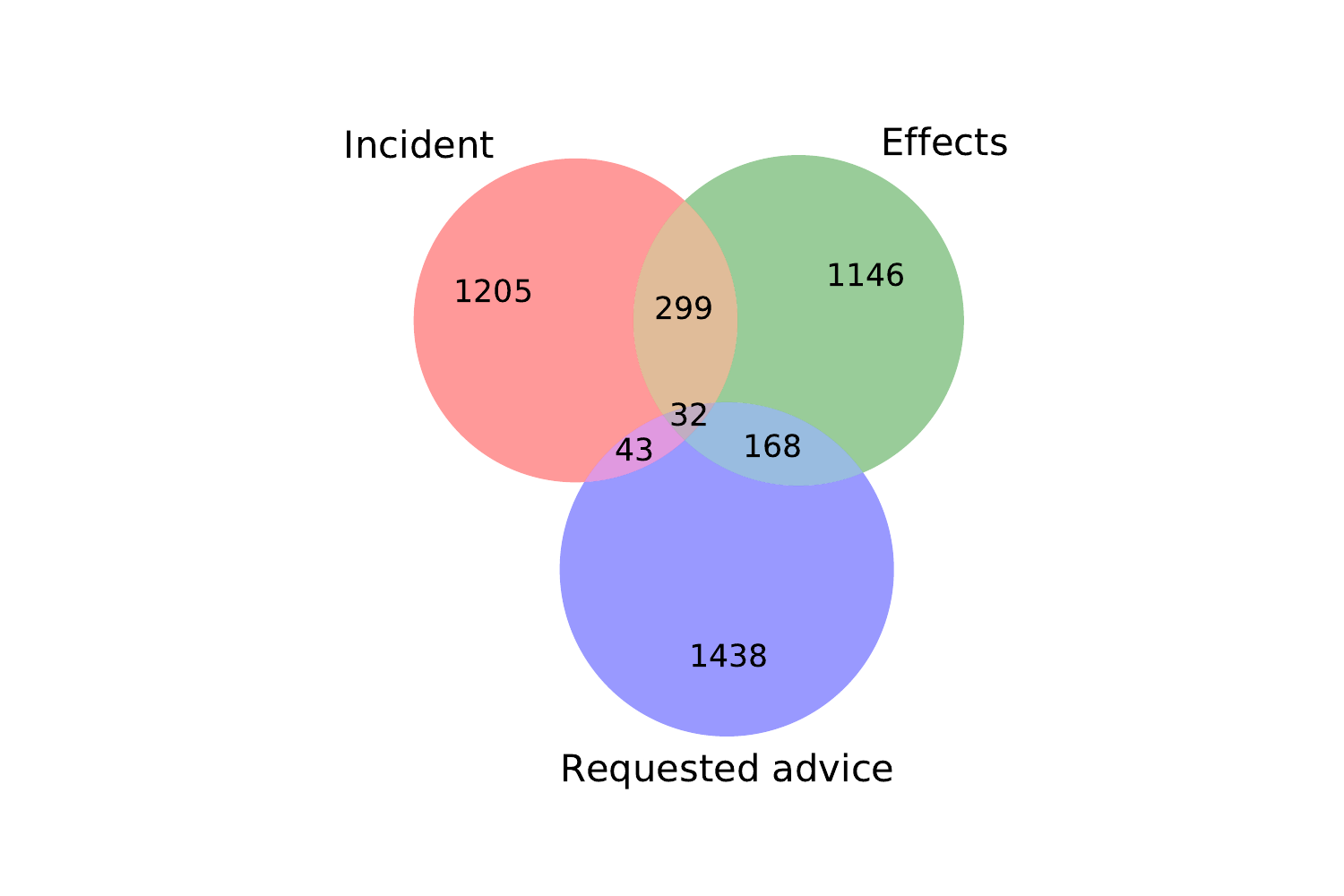}
    \caption{Venn diagram showing the distribution of sentences across the three categories.}
    \label{fig:venndistribution}
\end{figure}

In \corpus, there are \np{4331} (\np{48.4}\%) sentences that belong to at least one category, and \np{4616} (\np{51.6}\%) others. Figure~\ref{fig:venndistribution} shows the Venn distribution of \np{4331} sentences among three categories.

Finally, we trained XLNet on \corpus which is used to extract sentences from long posts. Over ten cross validation of \corpus, the model achieved \np{0.82} macro F1 score (\np{0.78} for incident, \np{0.79} for effects, and \np{0.89} for requested advice), \np{0.86} macro recall (\np{0.82} for incident, \np{0.83} for effects, and \np{0.92} for requested advice), and \np{0.78} macro precision (\np{0.74} for incident, \np{0.76} for effects, and \np{0.85} for requested advice).

\begin{figure*}[!htb]
\minipage{0.32\textwidth}
  \includegraphics[width=\linewidth]{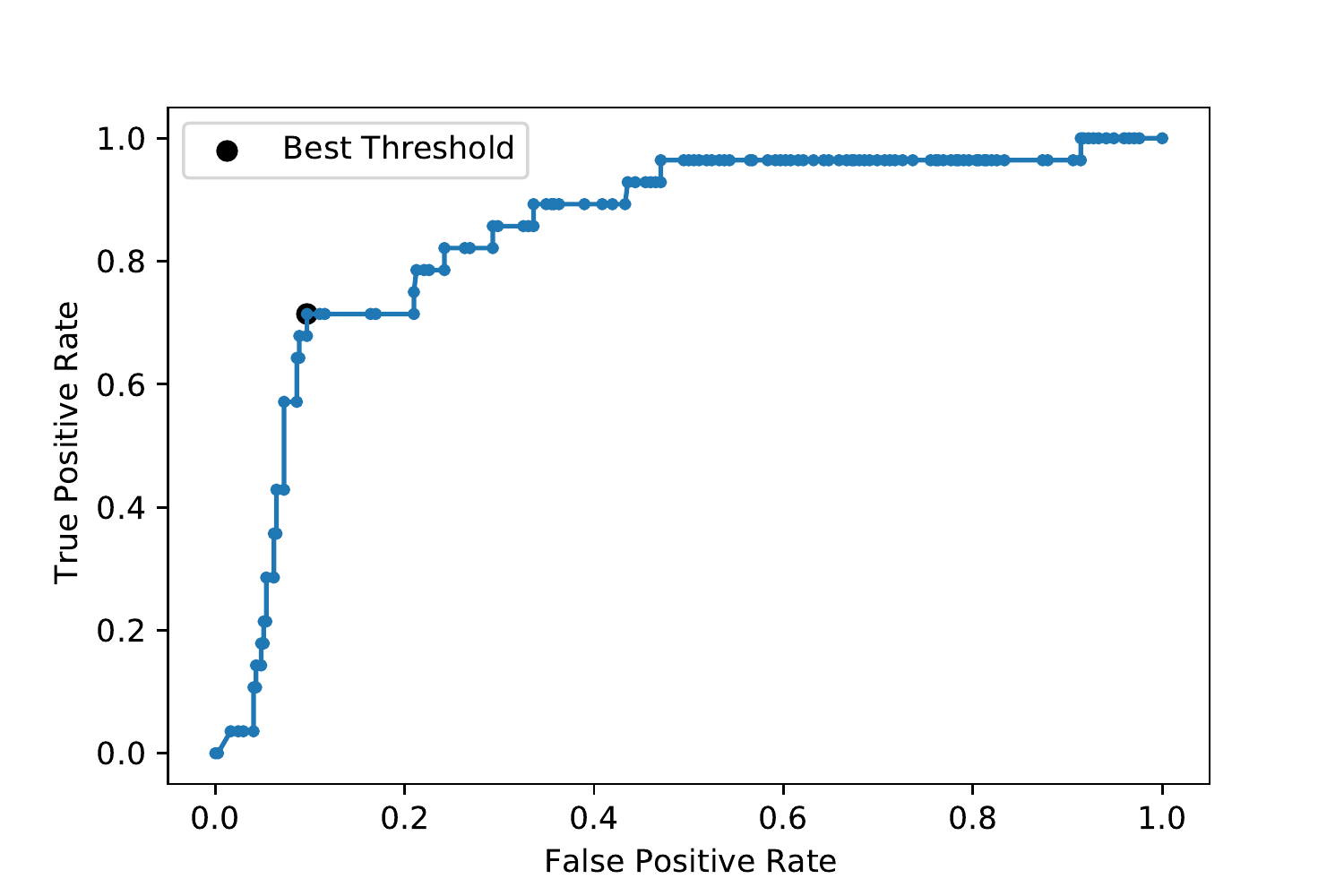}
  \caption{ROC curve showing true positive and false positive rates, while considering misclassified incident sentences under positive class. The area under the curve is 0.84. The best threshold is 0.038177.}\label{fig:rocincident}
\endminipage\hfill
\minipage{0.32\textwidth}
  \includegraphics[width=\linewidth]{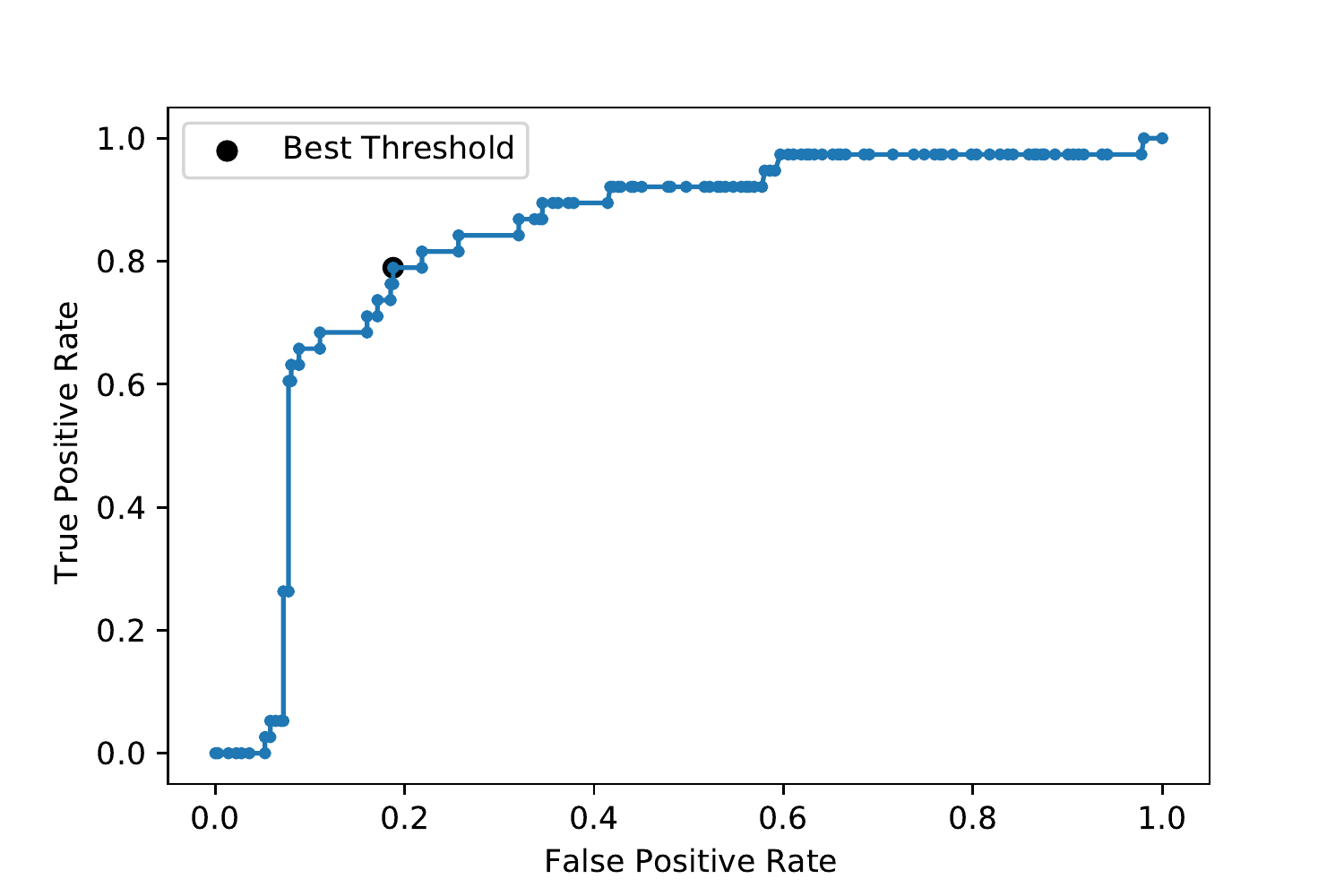}
  \caption{ROC curve showing true positive and false positive rates, while considering misclassified effects sentences under positive class. The area under the curve is 0.83. The best threshold is 0.008476. }\label{fig:roceffects}
\endminipage\hfill
\minipage{0.32\textwidth}%
  \includegraphics[width=\linewidth]{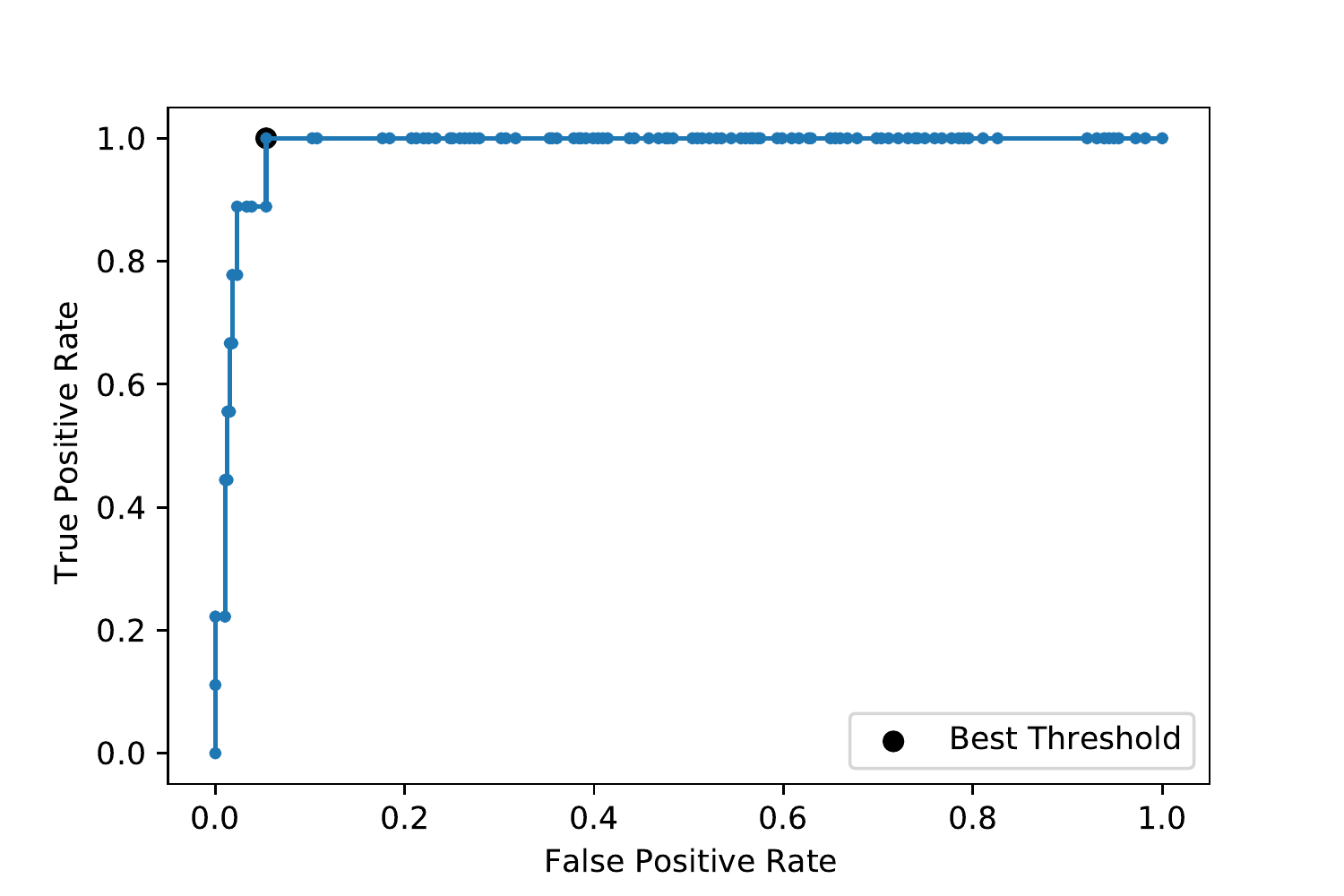}
  \caption{ROC curve showing true positive and false positive rates, while considering misclassified requested-advice sentences under positive class. The area under the curve is 0.98. The best threshold is 0.007874.}\label{fig:rocadvice}
\endminipage
\end{figure*}

\subsection{Selecting Query Method}
\label{sec:querying}
Uncertainty sampling \citep{Leastconfidence2005,Entropy1995}, a widely used querying method, finds uncertain predictions based on the model's prediction probability on the set U. Such uncertain data points are queried for manual labeling. However, uncertainty sampling methods (such as least confidence and entropy) did not work in our case. This is because in the first active learning cycle, model M (XLNet trained on \np{5947} sentences; Section~\ref{sec:approach}) predicted low probabilities on the sentences without any keywords (set U). We validated this by predicting on \np{100} such sentences, where the mean prediction probability was \np{0.08} (deviation= \np{0.23}) for incident category, \np{0.10} (deviation= \np{0.27}) for effects, and \np{0.05} (deviation= \np{0.21}) for requested advice. Due to most of the probabilities being low, uncertainty sampling methods (such as least confidence and entropy) could not discriminate between misclassified and other sentences.

To select an appropriate query method, we found a threshold on the prediction probability, using which we could query from U. To find that threshold, we used a set U', another random sample of \np{500} sentences not having any keywords. On U', we plotted the Receiver Operating Characteristic (ROC) curve and computed Youden's J-statistic \cite{Jstats1950} as described below. Since this query method was selected during the first active learning cycle, the model M referred below is XLNet trained on \np{5947} sentences (Section~\ref{sec:approach}).

\begin{enumerate}
\item The first author labeled the set U'.
On \np{5947} initially labeled sentences (Section~\ref{sec:labeleddata}), we achieved substantial agreement for the incident and effects category and almost perfect agreement for the requested advice. Hence, we assumed that all the annotators (three authors of this paper) stick to the same labeling definitions and used one of the annotators (the first author) for this small task.  

\item The model M made predictions on the set U'.

\item We split the set U' into a set of \np{400} sentences (set V) and another set containing remaining \np{100} sentences (set T).
    
\item We leveraged the set V to fine-tune the threshold.
For each category, we considered the misclassified sentences in set V and found a threshold (on the predictions' probability) that could retrieve them. We found that out of \np{400} sentences, model M  misclassified 28 sentences for the incident category, 38 for effects, and 9 for the requested advice category. Since we needed to retrieve these sentences, for each category, we considered misclassified sentences under positive class and others under negative class, while plotting  true positive and false positive rates in ROC. Figures~\ref{fig:rocincident}-~\ref{fig:rocadvice} show ROC and the area under the curve for each category. From each ROC curve, we found the best threshold (using Youden's J-statistic \cite{Jstats1950}) that maximized recall (for positive class) and minimized false positive rate. 

For the incident category, we found 0.038177 as the threshold, above or equal to which sentences can be queried. Similarly, we found a threshold of 0.008476 for the effects category and 0.007874 for the requested advice category. We also tried combining these three thresholds into a single threshold but that did not query more misclassified sentences than the individual threshold case. 
    
\item For each category, to ensure that we did not miss out on the misclassified sentences, we also queried 30 sentences below the threshold for every \np{100} predictions. For example, the model M predicted on the set V which has 400 sentences (4 times 100), we also queried 4*30 = 120 sentences below the threshold for each category.  

\emph{To sum up, our query method is: for each category, query (i) sentences with prediction probability above or equal to the threshold, and (ii) \np{30} sentences below the threshold for every \np{100} predictions.}

Using the above query method, we could query the following number of misclassified sentences from V: 25 (89.28\%) of 28 for the incident, 35 (92.10\%) of 38 for effects, and 9 (100\%) of 9 for requested advice. Since set V was used for fine-tuning, we also tested our query method on the unseen set T.

\item We leveraged the set T to test our query method. In the set T, M misclassified 10 sentences in the incident category, 4 in the effects, and 3 in the requested advice. Our query method retrieved 9 (90\%) misclassified incident sentences and all misclassified cases (100\%) in the other two categories. 



\end{enumerate}

For each time the active learning cycle was repeated (discussed in Section~\ref{sec:remainingal}), we used the above query method to retrieve potential misclassified sentences from U and manually labeled retrieved sentences. For manual labeling, each of the three annotators (authors of this paper) labeled retrieved sentences for a category.

\section{Qualitative Analysis}
\label{sec:outputanalysis}
We applied the final XLNet model (trained on \corpus as described in Section~\ref{sec:remainingal}) on random \np{20} posts (containing at least a thousand characters) and followed the below steps for each post. 

First, we split the post into multiple sentences, using the sentence tokenizer of the NLTK library.\footnote{\url{https://www.nltk.org/api/nltk.tokenize.html}}. Second, we provided all the sentences as input to model M and arranged the extracted sentences in the order they were present in the post. Third, along with the post title, we read the extracted sentences in the arranged order and checked if extracted sentences are coherent to understand the incident, effects, and requested advice. For \np{17} out of \np{20} posts, the extracted text was coherent. 

Further, we divide 17 posts and their extracted text among three annotators (the same authors A\textsubscript{1}, A\textsubscript{2}, and A\textsubscript{3}) such that each post was read by one annotator and its extracted text was read by the other. Each annotator was asked to construct a supportive or advice-offering response based on details present in the given text. Providing such responses is one kind of help to the survivor \cite{Helpfulresponse2019,Supportseeking2016}. For each post, the first author analyzed the difference between the response to the post (R\textsubscript{p}) and the response to the extracted text (R\textsubscript{e}). Only in \np{1} of \np{17} cases, a crucial detail (about the survivor's situation) was missed by R\textsubscript{e} that was part of R\textsubscript{p}. This was because that detail was also missing from the extracted text. However, for \np{16} of {17} cases, R\textsubscript{e} did not miss out any crucial details that were part of R\textsubscript{p}. Our model can potentially be used to understand the essential details (without reading long posts) and construct a helpful response based on the extracted text. In turn, this can speed up the process of providing help on a large scale.

\section{Psycholinguistic Analysis}
\label{sec:psycholinguistic}
LIWC-22 toolkit \cite{LIWC2022} contains 100 in-build dictionaries, where each dictionary consisted of lexicons, emoticons, and other verbal constructs to identify the psychological aspects from a post. We applied LIWC-22 on \corpus dataset and compared the sentences of the three categories on two types of scores: (i) summary and (ii) affect scores. We show our analysis below.

\subsection{LIWC-22 Summary Analysis}
\begin{figure}[!htb]
    \noindent\includegraphics[width=\columnwidth]{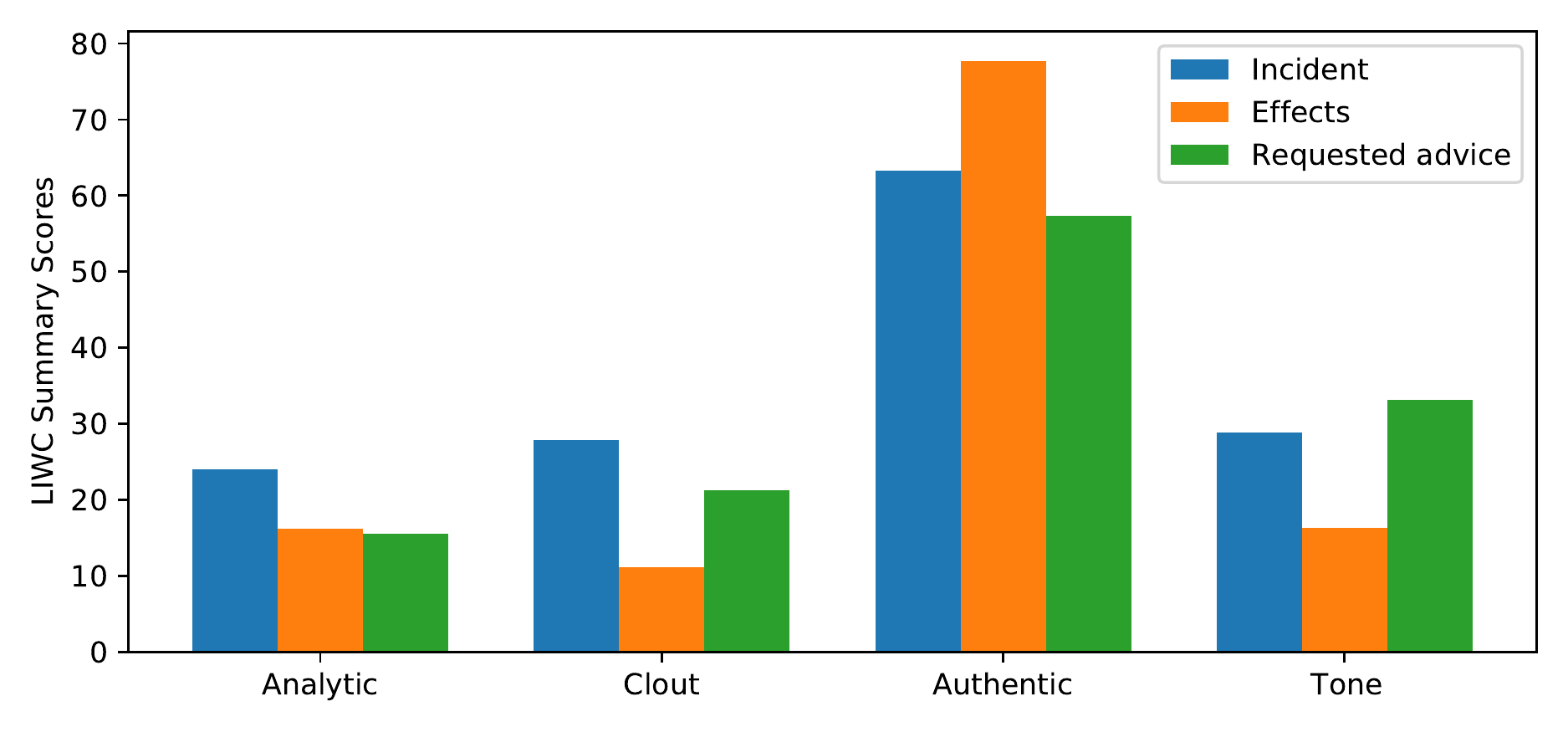}
    \caption{LIWC average summary scores for the incident, effects, and requested advice categories. Analytic and clout scores are low for each category. The low trend is also present in tone, indicating a negative tone in all three categories. However, high scores for authenticity indicate that the survivors openly share their experiences, especially through effects sentences.}
    \label{fig:summaryscore}
\end{figure}

For a sentence, LIWC-22 \cite{LIWC2022} yielded four types of summary scores: \fsl{analytic} (depicting analytical and logical thinking patterns), \fsl{clout} (depicting social status, confidence, or leadership), \fsl{authentic} (depicting how much people reveal about themselves, without any filter), and tone (depicting the emotional tone. The lower the score, the more negative the tone). On \corpus, we computed these summary scores (using the LIWC-22 toolkit) for each sentence that belongs to the incident, effects, and advice categories. Further, for each category, we averaged these scores. Figure~\ref{fig:summaryscore} shows category-wise average summary scores. \textbf{All three categories of sentences have low average scores for analytic and clout. This indicates less leadership, confidence, and logical thinking patterns in all such sentences. The same low trend is visible for the tone variable, meaning that all three categories of sentences possess negative tone}. Moreover, the most negative tone is observed in the effects category, leading to the lowest tone score.

However, the three categories have high average scores for authenticity. \textbf{We deduced that the survivors share their MeToo experiences without any filter. They are open to reveal about themselves, especially through the effects sentences (having the highest authenticity score)}. Through effects sentences, survivors reveal their feelings and emotions, which can be the reason for the highest authenticity score. Some examples of effects sentences having high authenticity include: \emph{``I feel worthless,'' ``I'm livid,'' and ``I'm scared.''}

\subsection{LIWC-22 Affect Analysis}
\begin{figure}[!htb]
    \noindent\includegraphics[width=\columnwidth]{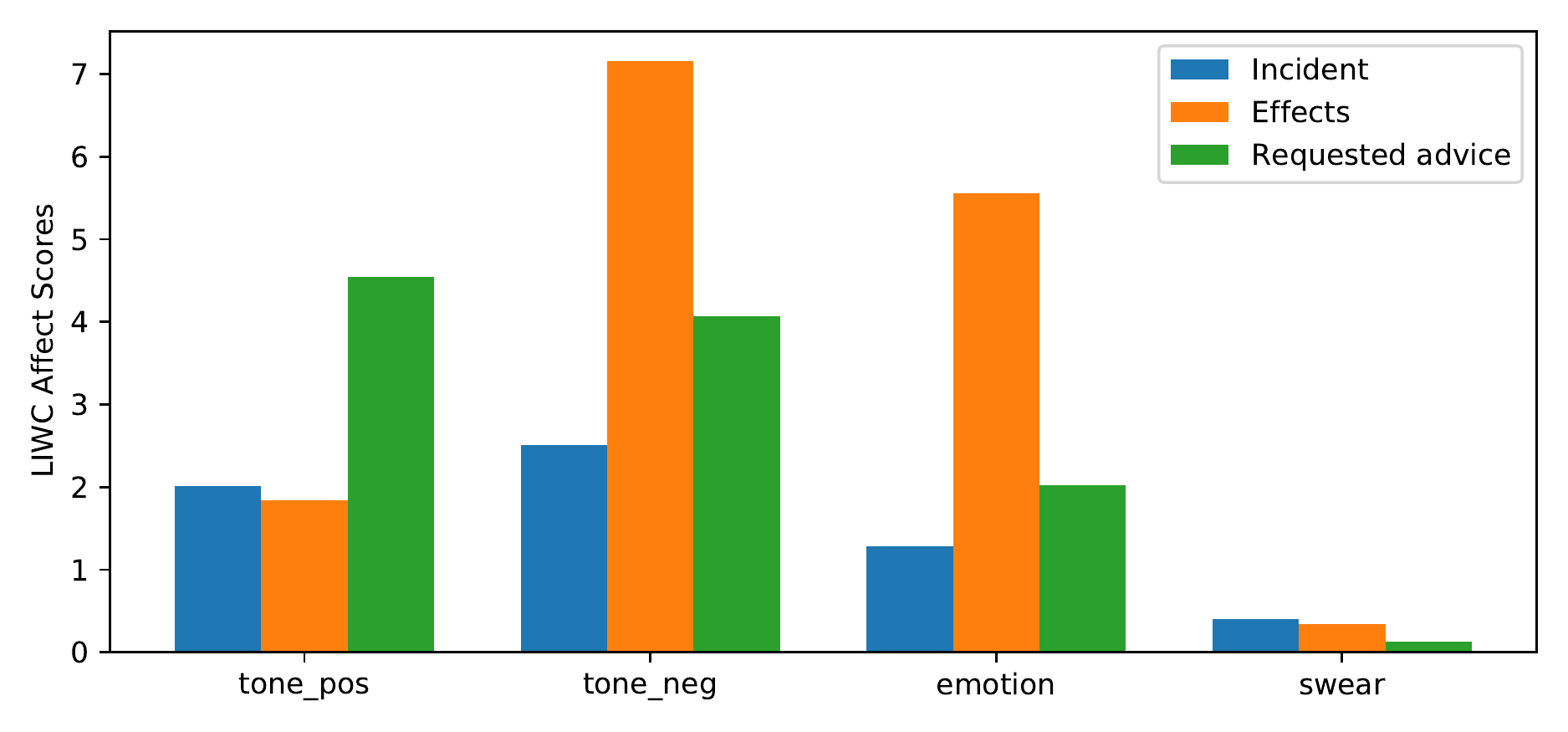}
    \caption{LIWC average affect scores. Effects sentences are more negative and emotional than incidents and requested advice. In terms of a positive tone, sentences requesting advice are ahead of incidents and effects. In general, all three categories of sentences don't contain swear words.}
    \label{fig:affectscore1}
\end{figure}

For a sentence, LIWC-22 \cite{LIWC2022} yielded four types of affect scores: \fsl{tone\_pos} (positive tone), \fsl{tone\_neg} (negative tone), \fsl{emotion}, and \fsl{swear}. Figure~\ref{fig:affectscore1} shows average affect scores for all three categories. It is evident that \textbf{effects sentences are more negative} ($\mu=7.16$) \textbf{and emotional} ($\mu=5.56$) \textbf{than incident} (tone\_neg $\mu=2.50$ and emotion $\mu=1.27$) \textbf{and requested advice} (tone\_neg $\mu=4.06$ and emotion $\mu=2.01$). Also, note that requested-advice sentences show a more negative tone than incident sentences. In such negative and advice seeking sentences, 
the survivors sometimes blame themselves and seek validation from others. For example, ``Is it my fault for drinking too much?''. Despite such negative cases, the same category shows many other positive tone cases. As a result, \textbf{in terms of positive tone, sentences requesting advice} ($\mu=4.54$) \textbf{are ahead of incident} ($\mu=2.01$) \textbf{and effects sentences} ($\mu=1.84$). Moreover, we found that all three categories, in general, don't contain swear words, as depicted by their low swear scores.

As effects sentences are more emotional than the other two categories, we delved into what types of emotions are reflected by effects sentences. For a sentence, LIWC-22 yielded four types of emotional scores: \fsl{emo\_pos} (positive emotion), \fsl{emo\_anx} (anxiety), \fsl{emo\_anger} (anger), and \fsl{emo\_sad} (sadness). We found that \textbf{effects sentences show more anxiety ($\mu=1.53$), followed by sadness ($\mu=0.70$), anger ($\mu=0.43$), and positive emotion ($\mu=0.39$)}.

\section{Related Work}
\label{sec:relatedwork}

There has been extensive research in analyzing MeToo posts and finding useful insights \citep{Metoolens2018,Metooma2020,DidNotReport2020, Contextualaffective2019,DiscourseAnalysis2020}. However, only a few studies have looked MeToo experiences from classification perpective. \citet{Safecity2018} leverage the MeToo experiences posted on the SafeCity website\footnote{\url{https://www.safecity.in/}}, an online forum to report sexual harassment. They collect \np{9892} MeToo experiences that convey one of the three types of harassment: (i) groping or touching, (ii) staring or ogling, and (iii) commenting. Further, they train a deep neural network to identify the type of harassment experienced by the survivor. \citet{Quantumsafecity2019} improve the performance of this classification by proposing a quantum-inspired density matrix encoder.
\citet{Storyclassification2019} leverage the same dataset and annotate it for attributes such as the abuser's age (below 30 or older), the abuser's relation with the survivor (for example, relative or teacher), location of harassment (for example, park or street). They propose a framework to identify these attributes from a MeToo experience. \citet{MeTooMaastricht2020} also leverage the SafeCity dataset and build a chatbot system to help survivors. The SafeCity dataset contains concise experiences (typically 3-4 sentences long) and is unfit to extract sentences in our case. 

Moreover, \citet{SV-Tracking2020} train a model on \np{520761} \#MeToo hashtag tweets to identify tweet level attributes, such as the category of sexual violence reported, the survivor's identity (tweeter or not), the survivor's gender. They also achieve 80.4\% precision and 83.4\% recall in identifying sexual violence reports. \citet{Youtoo2019} label \np{5119} tweets for types: (i) disclosure and (ii) nondisclosure. The tweets that include a survivor's personal experience are annotated as disclosure and others as non-disclosure. Out of \np{5119}, they find \np{1126} (22\%) tweets under disclosure category. Moreover, they propose a language model to classify tweets into two types. Moreover, other studies such as \citet{Silence-breakers2018} and \citet{Metoo-Disclosure2019} also focus on similar classification tasks. All these works perform classification tasks on each MeToo post. However, our work focuses on sentence-level extraction of sexual harassment incident, its effects on the survivor, and requested advice. 

Traditional text summarization works \citep{SwapNet2018,Cheng-lapata-neural-2016,See-etal-2017,Aspectsummarizer2011,Coherentsummary2012} are trained or evaluated on other domain specific datasets such as news datasets but are not built for MeToo context. Our work is the first attempt to extract text from long MeToo posts to the best of our knowledge.

\section{Discussion}
\label{sec:conclusion}
We now discuss our conclusion, our work's limitations, and possible future directions.

\subsection{Conclusion}
The survivors of sexual harassment frequently share their long MeToo posts on subreddits. Using the active learning approach, we trained XLNet model to extract sentences describing (i) the sexual harassment incident, (ii) the effects on the survivor, and (iii) the requested advice, from such posts. We also curated \corpus, a dataset of \np{8947} sentences labeled for the three categories, and conducted psycholinguistic analysis on it. On ten-fold cross-validation of \corpus, our model achieved a macro F1 score of \np{0.82}.
The sentences extracted by our model can help a prospective helper understand essential details without having to read the entire post. As a result, it can potentially speed up the process of providing help to the survivors.

\subsection{Limitations and Future Work}
Our work suffers from some limitations, and a few of them also motivate future directions of improvement. First, sometimes the extracted sentences may not be coherent or miss some details about the survivor's situation. That's why we don't claim our model to be a summarization tool. However, according to our analysis in Section~\ref{sec:outputanalysis}, non coherent cases and the cases requiring details beyond the extracted text are only a few (4 of 20). In the future, our work could be extended to extract other important sentences which can summarize the whole post. Second, our model is trained on the sentences scraped from only three subreddits. We expect the nature of sentences in \corpus to be similar to sentences present on other MeToo-related subreddits. However, we plan to fine-tune the model before applying it on the other subreddits. 

We can also extend our work to generate an automated response based on the extracted incidents, effects, and requested advice. The automated response after slight corrections by human interventions can offer support and advice to the survivor. Moreover, the similarity between the advice-seeking sentences and users' responses can assess how relevant each response is. This way platform will be able to show highly relevant responses above the less helpful ones.

\section{Broader Perspective and Ethical Considerations}
\label{sec:ethics}
Although we extracted text from long posts to help survivors, we acknowledge some limitations and possible misinterpretations that may occur, especially with the data on such a sensitive topic. We discuss them below.

\begin{enumerate}
\item \textbf{Consent:} Our data was scraped from the public Reddit posts. Hence, we did not take the consent of the survivors writing such posts. Moreover, as described by \citet{Youtoo2019}, some survivors may get uncomfortable if they are reached out for consent. 
    
\item \textbf{Anonymity:} We did not save survivors' personal information such as usernames, or users' history of posts. 
For the example sentences presented in this paper, we also removed potentially identifying information, such as the survivor's age, job title, and location. Moreover, we paraphrased the example MeToo post. We don't plan to release \corpus publicly.

\item \textbf{Labeling disturbing text:} The sentences from MeToo posts can be disturbing to read, especially for people who have gone through a similar experience. Therefore, we didn't hire crowd workers or volunteers to for any labeling task. Instead, the three authors of this paper did it.

\item \textbf{Potential misinterpretation:} We were extremely aware of the sensitivity of this research before labeling sentences. However, we may have misinterpreted some MeToo experiences. That's why we don't claim that our labeling is fully accurate.

\end{enumerate}

\bibliographystyle{IEEEtranSN} 
\bibliography{Vaibhav}

\end{document}